\documentclass{article}

\usepackage{microtype}
\usepackage{graphicx}
\usepackage{subfigure}
\usepackage{booktabs} 

\usepackage{hyperref}

\usepackage[accepted]{icml2023}

\usepackage{amsmath}
\usepackage{amssymb}
\usepackage{mathtools}
\usepackage{amsthm}

\usepackage[capitalize,noabbrev]{cleveref}

\theoremstyle{plain}
\newtheorem{theorem}{Theorem}[section]

\newtheorem{lemma}[theorem]{Lemma}

\theoremstyle{definition}

\theoremstyle{remark}

\usepackage[textsize=tiny]{todonotes}
\usepackage{booktabs}
\usepackage{graphicx}
\usepackage{multirow}
\icmltitlerunning{Long Range Continuous-time Conterfactual Outcome Prediction using Structured State Space Model}

\begin{document}

\twocolumn[
\icmltitle{Counterfactual Outcome Prediction using Structured State Space Model}

\icmlsetsymbol{equal}{*}

\begin{icmlauthorlist}
\icmlauthor{Vishal Purohit}{yyy}
\end{icmlauthorlist}

\icmlaffiliation{yyy}{Purdue University}

\icmlcorrespondingauthor{Vishal Purohit}{purohitv@purdue.edu}

\icmlkeywords{conterfactual outcomes, state space model, continuous time modeling}
\vskip 0.3in
]

\printAffiliationsAndNotice 

\begin{abstract}
Counterfactual outcome prediction in longitudinal data has recently gained attention due to its potential applications in healthcare and social sciences. In this paper, we explore the use of the state space model, a popular sequence model, for this task. Specifically, we compare the performance of two models: Treatment Effect Neural Controlled Differential Equation (TE-CDE) and structured state space model (S4Model). While TE-CDE uses controlled differential equations to address time-dependent confounding, it suffers from optimization issues and slow training. In contrast, S4Model is more efficient at modeling long-range dependencies and easier to train. We evaluate the models on a simulated lung tumor growth dataset and find that S4Model outperforms TE-CDE with 1.63x reduction in per epoch training time and 10x better normalized mean squared error. Additionally, S4Model is more stable during training and less sensitive to weight initialization than TE-CDE. Our results suggest that the state space model may be a promising approach for counterfactual outcome prediction in longitudinal data, with S4Model offering a more efficient and effective alternative to TE-CDE.
\end{abstract}

\section{Introduction}
Healthcare systems are dynamic and continuously changing in order to improve care and treatments for patients. In order to accomplish this goal, healthcare providers rely on accurate and dependable tools that can predict patient outcomes and provide guidance for treatment decisions. Standard clinical risk prediction models are commonly used to forecast an individual's risk of an outcome based on their observed characteristics. However, these models are frequently created using data from a population in which patients follow a mix of treatment strategies. This makes them unsuitable for informing treatment decisions. Counterfactual prediction models offer an alternative approach that estimates a person's risk of an outcome if they were to follow a specific treatment pattern, taking into account other patient characteristics that predict the outcome. Despite their potential advantages, the development of decision models that predict counterfactual outcomes presents several challenges. These challenges include determining the appropriate methods for identifying and adjusting for confounding factors, accounting for time-varying treatments and outcomes, and incorporating complex interactions between different patient characteristics. Additionally, the high dimensionality of healthcare data and the need to balance model complexity with interpretability and generalizability further complicates the creation of accurate and reliable counterfactual prediction models in healthcare. Nonetheless, the development of such models has the potential to greatly enhance healthcare decision-making and improve patient outcomes.

When dealing with longitudinal data, the observed data can occur at irregular intervals, adding an additional layer of complexity that must be accounted for by the algorithm. In addition to the challenge of irregular time-series data, we also face the problem of \textit{time-dependent confounding}. Confounding variables are factors that are associated with both the exposure and the outcome of interest, and they can introduce bias into causal inference if not properly accounted for. In the context of cancer treatment, the history of patients' covariates and their response to past treatments can be important confounding variables that affect the choice of future treatments. Failing to account for these variables can lead to biased estimates of the causal effect of a particular treatment on patient outcomes. Addressing this issue requires the use of various statistical methods, such as propensity score matching or instrumental variables, to adjust for confounding variables and obtain more accurate estimates of causal effects. However, identifying the appropriate confounding variables to include in the analysis and the optimal method for adjusting for them can be challenging. Moreover, the availability and quality of data for these variables may vary across patients, making it difficult to achieve a balanced and fair comparison between treatment groups. Thus, it is crucial to carefully consider the potential sources of bias and confounding in longitudinal data analysis and apply appropriate methods to minimize their impact on causal inference. Additionally, developing novel and innovative statistical methods that can handle irregular longitudinal data and time-dependent confounding is an important area of ongoing research in the field of causal inference.

Specifically, time-dependent confounding and distribution shifts pose unique challenges in causal inference over time, which are not typically encountered in standard time-series analysis. Conventional time-series models do not account for the bias introduced by time-varying confounders, and therefore may not be appropriate for analyzing observational data in which there is a potential for confounding. To address this issue, researchers may need to use more sophisticated causal inference methods that can account for time-varying confounding and distribution shift, such as time-varying treatment models or g-methods. These methods allow for the estimation of the causal effect of a treatment or intervention over time, while also accounting for the potential biases introduced by confounding factors that change over time.

Our study differs from TE-CDE \cite{tecde}, which uses a continuous-time model for estimating treatment effects over time. We use a general sequence-to-sequence model that does not rely on continuity. Despite this difference, our model significantly outperforms TE-CDE, achieving up to 100 times better results in some cases. For critical applications such as counterfactual outcome estimation in a medical setting, it is essential to have a model that can provide reliable and accurate estimates of treatment effects. Our approach may not have the same theoretical soundness as TE-CDE, but it has practical advantages in terms of computational efficiency and scalability. However, we need further investigation to understand the factors that affect the performance gap between our model and TE-CDE, and to explore the strengths and weaknesses of each approach in different scenarios. We also aim to develop new and innovative models that can combine the best features of both approaches.
\section{Related Work}
This paper primarily deals with counterfactual outcome estimation in irregularly sampled settings with time-dependent confounding. We briefly outline the details of the method that is closest to our problem. 
\subsection{Continuous-Time Modeling of Counterfactual Outcomes Using Neural
Controlled Differential Equations}
The ability to estimate counterfactual outcomes over time has immense potential to revolutionize personalized healthcare by enabling decision-makers to answer "what-if" questions. However, current causal inference approaches are limited in their ability to handle irregularly sampled data, which is a common occurrence in practice. To address this issue, \cite{tecde} propose a novel approach called Treatment Effect Neural Controlled Differential Equation (TE-CDE), which models the underlying continuous-time process explicitly using controlled differential equations. This allows for the potential outcomes to be evaluated at any time point, even in the presence of irregularly sampled data. TE-CDE incorporates adversarial training to account for time-dependent confounding, which is a crucial challenge in longitudinal settings not encountered in conventional time-series analysis. The performance is evaluated on data generated by a lung tumour growth model that reflects a range of clinical scenarios with irregular sampling. 

\subsection{SyncTwin: Treatment Effect Estimation with Longitudinal Outcomes \cite{synctwin} and SurvITE: Learning Heterogeneous Treatment Effects
from Time-to-Event Data \cite{survite}}
Though not related directly to our problem, we briefly explain these two papers that deal with treatment effect estimation.  SyncTwin is proposed for estimating causal treatment effects using electronic health records. SyncTwin learns a patient-specific time-constant representation from pre-treatment observations to issue counterfactual predictions. SyncTwin is demonstrated to be usable in real-world EHR and successfully reproduced the findings of a randomized controlled clinical trial using observational data. On the other hand SurvITE discussed the problem of inferring treatment effects from time-to-event data, which involves estimating treatment effects on both instantaneous risk and survival probabilities.
\section{Problem Formulation}
We follow the same problem formulation as in work \cite{tecde},. For completeness, we summarize the problem formulation. Let $n$ be the number of individuals whose covariates are collected over time [0,T]. Let $\textbf{X}: [0, T] \rightarrow \mathbb{R}^{d}$. The covariate path over time $[0,T]$ of each patient is $x_i \in \mathbb{R}^d$. Let $A$ define a treatment process over period of time $[0,T]$ where $a_i \in \{0,1\}$.The output $Y$ is dependent on treatment $A$ and patient covariates $X$. In our paper, we examine the tumour size growth hence $y_i \in \mathbb{R}$ and $\textbf{Y} : [0,T] \rightarrow \mathbb{R}$. We explicitly define time $\mathcal{T}$ to be a record of times at which the observation from the healthcare worker is made. For $i^{th}$ patient the record of time is denoted by $t_i \in \mathcal{T}$.
If $\text{S = \{$X_s, A_s, Y_s$\}}$ represents a set of observations over a period of time $[0, s]$. In longitudinal studies, the observed data for each patient may be collected at irregular intervals due to various reasons, such as missed appointments, treatment schedules, or patient and healthcare worker preferences. This irregularity adds an additional layer of complexity to the analysis, as the algorithm must be able to handle missing data and account for the variability in observation times. In our study, we aim to provide estimates of treatment effects at a specific time point beyond the initial observation interval [0, s], denoted as $t_{i}^{'}$. To accomplish this goal, we make certain assumptions and use a specific formulation that is described in more detail in \cite{tecde}. By addressing the challenges posed by irregularly sampled longitudinal data, we can obtain more accurate and reliable estimates of treatment effects, which can help inform clinical decision-making and improve patient outcomes.  
\vspace{-3mm}
\section{S4Model : Structured State Space Model}
 The state space model (SSM) is a foundational scientific model mainly used in fields such as control theory and computational neuroscience. A state space model is defined as follows,
\begin{align}
     x^{'}(t) = \textbf{A}x(t) + \textbf{B}u(t),\\
     y(t) = \textbf{C}x(t) + \textbf{D}u(t).
 \end{align}
In the field of control theory, the state space model (SSM) is a mathematical framework used to describe the behavior of dynamic systems. It consists of a set of differential equations that relate the system's state variables to its inputs and outputs. The state vector $x(t)$ is an N-dimensional vector that represents the internal state of the system at time t, while the output vector $y(t)$ is a vector that represents the system's measurable outputs. The input signal $u(t)$ is a one-dimensional vector that represents the inputs to the system.

The state, input, output, and feedforward matrices $\textbf{A, B, C, D}$ are used to define the relationships between the state, input, and output vectors. In recent work, SSM has been treated as a black box representation of the sequence model where matrices $\textbf{A, B, C, D}$ are learned using gradient descent. To encode long sequences, HiPPO theory of continuous time memorization has been leveraged, which specifies a class of certain matrices $\textbf{A} \in \mathbb{R}^{N \times N}$ that allows the state $x(t)$ to memorize the history of the input $u(t)$.

Linear State Space Models (LSSL) \cite{lssl} have been studied extensively in the literature and have been shown to have multiple interpretations as an ordinary differential equation, a recurrent model, and a convolution. This paper focuses on structured space models \cite{structured}, which are easier to train and offer superior performance in relation to other sequence models. For more details, readers can refer to the rich literature on state space models.

Under the discrete-time view of LSSL, given an input sequence $(u_1, u_2, ...)$ discretized by time step of $\Delta t$, the resulting discretized version can be obtained with the help of bilinear transform that converts matrix $\textbf{A}$ to $\Bar{\textbf{A}}$. The discrete SSM (assume $\textbf{D} = 0$ for simplicity) is given by,
\begin{align}\label{eq:eq_2}
    x_k = \Bar{\textbf{A}}x_{k-1} + \Bar{\textbf{B}}u_k,\\
     y_k = {\textbf{C}}x_k. \label{eq:eq_3}
\end{align}

Now the SSM in \cref{eq:eq_2} and \cref{eq:eq_3} is a mapping between input $u_k \rightarrow y_k$. The matrix $\Bar{\textbf{A}} = (\textbf{I} - \frac{\Delta}{2}.\textbf{A})^{-1}(\textbf{I} + \frac{\Delta}{2}.\textbf{A})$ and $\Bar{\textbf{B}} = (\textbf{I} - \frac{\Delta}{2}.\textbf{A})^{-1} . \Delta{\textbf{B}}$. The fundamental bottleneck in computing the discrete-time SSM \cref{eq:eq_2} that it involves repeated matrix
multiplication by \textbf{A}. For example, computing \cref{eq:eq_2} naively as in the LSSL involves L successive multiplications
by \textbf{A}, requiring $O(N^{2}L)$ operations and $O(NL)$ space. According to Lemma 3.2 in \cite{structured}, when \textbf{A} is structured then one can perform faster computation. For completeness, the lemma is as follows,
\begin{lemma}
(Diagonalization of \textbf{A})
Conjugation is an equivalence relation on SSMs $(\textbf{A}, \textbf{B}, \textbf{C}) \sim (\textbf{V}^{-1}\textbf{A}\textbf{V} ,\textbf{V}^{-1}\textbf{B}, \textbf{C}\textbf{V})$ and resulting SSM is, $\tilde{x}^{'} = \textbf{V}^{-1}\textbf{A}\textbf{V}x_{k-1} + \textbf{V}^{-1}\textbf{B}u_k$ and $y_k = \textbf{C}\textbf{V}x_k$ with $x = \textbf{V}\tilde{x}$
\end{lemma}
The diagonalization of HiPPO matrix \textbf{A} is unstable due to large values in matrix $\textbf{V}$ rendering the diagonalization numerically infeasible. \cite{structured} observed that HiPPO matrix can be decomposed as a sum of a normal and low-rank matrix.  With help of linear algebraic techniques \cite{structured} showed the following,
\begin{lemma}
 (Normal Plus Low Rank represenation (NLPR) of \textbf{A})   
 All HiPPO matrices have an NPLR representation
$\mathbf{A} = \mathbf{V}\mathbf{\Lambda}\mathbf{V}^{*} - \mathbf{P}\mathbf{Q}^{T}
= \mathbf{V}(\mathbf{\Lambda} - (\mathbf{V}^{*}\mathbf{P}) (\mathbf{V}^{*}\mathbf{Q})^{*})\mathbf{V}^{*}$ for unitary $\mathbf{V} \in \mathbb{C}^{N\times N}$ , diagonal $\mathbf{\Lambda}$, and low-rank factorization $\mathbf{P}$ , $\mathbf{Q} \in \mathbb{R}^{N\times r}$
\end{lemma}
Under the NLPR formulation over $\mathbb{R}$ the complexity of SSM reduces to $O(N)$ operations where $N$ is the state size.
\subsection{From S4 to Deep S4}
In order to initialize an SSM with \textbf{A} HiPPO matrix, we need to specify the matrix \textbf{A} with a specific initialization. This SSM is unitarily equivalent to some $(\mathbf{\Lambda} - \mathbf{P} \mathbf{Q}^{*}
, \mathbf{B}, \mathbf{C})$ for some diagonal $\mathbf{\Lambda}$ and vectors $\textbf{P} , \textbf{Q}, \textbf{B}$, $\textbf{C} \in \mathbb{C}^{N\times1}$. These matrices and vectors comprise S4's 5N trainable parameters.

S4 is a sequence model that defines a map from $\mathbb{R}^{L} \rightarrow \textbf{R}^{L}$, i.e., a 1-D sequence map. However, deep neural networks (DNNs) typically operate on feature maps of size H instead of 1. To handle multiple features, S4 defines c-independent copies of itself and then mixes the H features using a position-wise linear layer. This results in a total of $O(H^2)+O(HN)$ parameters per layer. Nonlinear activation functions are also inserted between these layers. 

\begin{figure*}[!ht]
   \includegraphics[width=\textwidth]{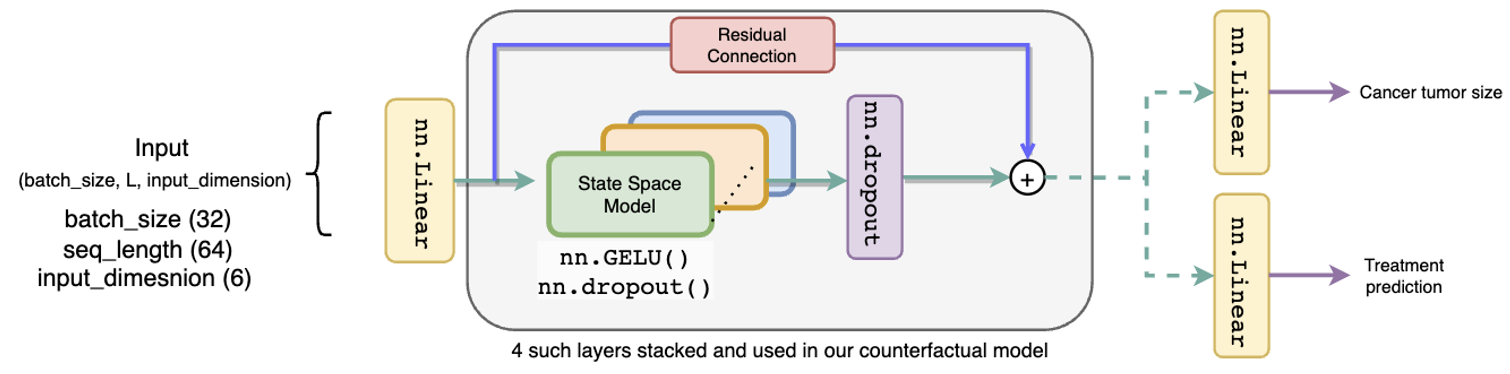}
   \caption{Diagram of Deep S4Model.}
   \label{fig:deep_s4}
\end{figure*}

Overall, S4 defines a sequence-to-sequence map of shape (batch size, sequence length, hidden dimension). It is worth noting that the core S4 module is a linear transformation, but the addition of non-linear transformations through the depth of the network makes the overall deep SSM non-linear. A diagram of S4model is show in Figure \ref{fig:deep_s4} and respective hyper-parameters are shown in Table \ref{tab:hypertab}.

It is important to note that S4 is designed for structured space models, which are easier to train and offer superior performance compared to other sequence models. In previous work on Linear State Space Models (LSSL) \cite{lssl}, they have been shown to have multiple interpretations, such as an ordinary differential equation, a recurrent model, and as convolution. Readers can refer to the rich literature on state space models \cite{lssl, structured} for further details. Thus, S4 is a deep structured space model that leverages the HiPPO theory of continuous time memorization to encode long sequences. By defining H-independent copies of itself and using position-wise linear layers to mix features, it can handle multiple features and achieve superior performance compared to other sequence models.

\subsection{Optimization Objective}
In our problem formulation, we have have a deep S4 model learning from $S = \{X_s, A_s, Y_s\}$ to predict $Y_{t^{'} > s}$ and $A_{t^{'} > s}$. Let the $f_{\phi}: \mathbb{R}^{c \times L} \rightarrow \mathbb{R}^{c \times L}$ represent the mapping function represented by $c$ S4 models and $\phi$ represent all the learnable parameters. let output $f_{\phi}$ be given by latent state $z$. This state is $z$ and is mapped to the treatment outcome (a) and lung tumour size (y) using two different linear transformations. The predicted outcome is over a period of time $[t, t_k]$ with observation times $(t_1,.....,t_k)$. The mean square error (MSE) of outcome prediction is defined as, 
\begin{equation}
    \mathcal{L}^{(y)} = \frac{1}{k} \Sigma_{i=1}^k (y_{t_i} -  \hat{y}_{t_i})^{2}.
\end{equation}
The treatment loss is calculated using cross-entropy loss, which is given by, 
\begin{equation}
    \mathcal{L}^{(a)} = -\frac{1}{k} \Sigma_{i=1}^k \bigg(a_{t_i} \log (\hat{a}_{t_i}) + (1 - a_{t_i} \log (1 - \hat{a}_{t_i})\bigg)
\end{equation}
We use a similar formulation as \cite{tecde}, to balance the representations and minimized the following loss function, 
\begin{equation}\label{eq:eq_7}
    \mathcal{L}_{total} = \frac{1}{n} \Sigma_{i=1}^n \bigg(\mathcal{L}_{i}^{(y)} - \mu \mathcal{L}_{i}^{(a)}\bigg),
\end{equation}
where $\mu$ is the hyper-parameter controlling the trade-off between treatment and outcome prediction. Unlike TE-CDE, we fix the value of $\mu = 0.5$ implying outcome prediction to be prioritized over treatment predictions.
\vspace{-3mm}
\section{Experiments}
In this section, we evaluate the ability of our proposed method to estimate counterfactual outcomes using observational data. In real-world scenarios, it is often impossible to obtain counterfactual outcomes, hence we use synthetic data to perform an empirical evaluation of our approach. To simulate the synthetic data, we utilize a simulation environment based on a Pharmacokinetic-Pharmacodynamic (PK-PD) model of lung cancer tumor growth \cite{tecde} This simulation framework allows us to compute counterfactual outcomes for any time point and treatment plan. However, to mimic the nature of real-world data, we simulate the irregularly sampled data using the Hawkes process.

Our evaluation consists of two main parts. Firstly, we evaluate the accuracy of our proposed method in estimating counterfactual outcomes for a given treatment plan. Secondly, we compare the performance of our approach with other state-of-the-art methods like TE-CDE for counterfactual estimation. We estimate the counterfactual outcomes using our proposed approach and calculate the mean squared error (MSE) between the estimated and simulated outcomes.

\subsection{Datasets and Architecture}
The PK-PD model used in our study has been previously described in \cite{tecde} Our simulation environment allows for the manipulation of three key parameters governing the generation of synthetic data. These parameters, $\gamma_c$ and $\gamma_r$, are responsible for controlling the degree of \textit{time-dependent confounding}, whereby higher values of $\gamma_{c,r}$ indicate that treatment assignment is influenced by tumour size. Cancer staging details and the use of the Hawkes process are elaborated in \cite{tecde}. Within the Hawkes process, $\kappa$ is used to control the intensity of data sampling between cancer stages. Specifically, higher stages of cancer result in more frequent sampling, thus reflecting some of the characteristics of real-world data that our algorithm may encounter. By manipulating these parameters within our simulation framework, we are able to generate synthetic data with known counterfactual outcomes, which serves as a robust testing ground for the evaluation of our proposed method.

The deep S4 model consists of 4 layers with the latent dimension of 256. The TE-CDE model uses a latent dimension of 128 for each of the two networks used for treatment and outcome prediction. Both models use a batch size 32 for training. In total, we have 10000 training datapoint, 1000 validation data points (used for hyper-parameter tuning) and 10,000 test data points. We use the validation data to find hyperparameters like  learning rate. The best learning rate for S4Model is 0.0005 for parameters \textbf{A, B, C} and 0.00002 for other parameters. We use learning rate of 0.00001 for TE-CDE as it was the best and most stable one. Both TE-CDE and S4Model are trained for 50 epochs on Nvidia A30 GPU with 24GB VRAM. 
\vspace{-3mm}
\subsection{Main Experiements}
\begin{figure*}[t]
\begin{center}
   \includegraphics[width=\textwidth]{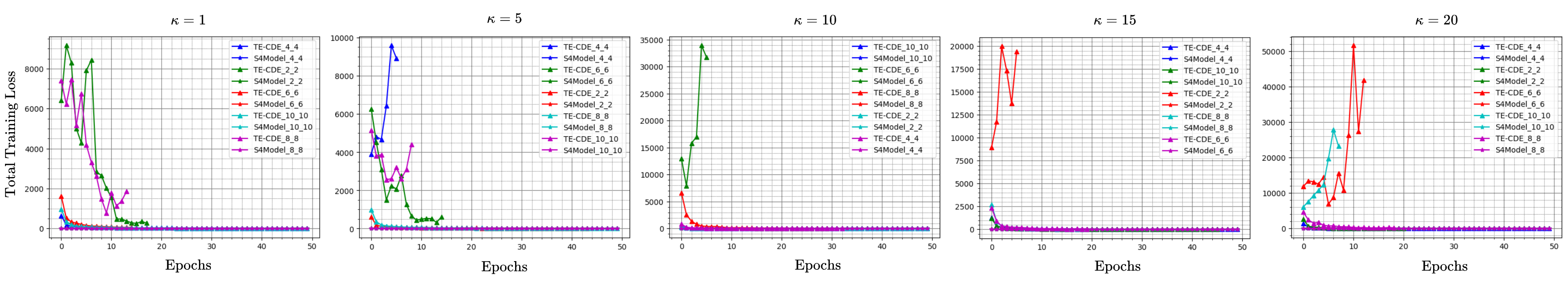}
   \vspace{-10mm}
   \caption{Total loss (cross entropy + MSE loss) v/s epochs for various $\kappa = \{1, 5, 10, 15, 20\}$. The values of $\gamma_{r,c} = \{2,4,6, 8, 10\}$. The training curves and respective configuration of $\gamma_{r,c}$ are shown as the last two numerical values in the plot legend.}
   \label{fig:fig_1}
   \end{center}
\end{figure*}

In \cref{tab:result_1} we show the side-by-side comparison of TE-CDE and S4Model performance on the test dataset. Here $\kappa = \{1,5, 10, 15, 20\}$, which shows difficult levels of sampling intensity of the Hawks process. Higher value of kappa implies sampling intensity is greater for more severe cancer stages. Different values of $\gamma_{r,c} = \{2, 4, 6, 8, 10\}$ control the \textit{time-dependent confounding}. Under sever time-confounding (higher values of $\gamma_{r,c})$ our method still outperforms TE-CDE by a significant margin. We observe that TE-CDE performs poorly in almost all combinations of $kappa$ and $\gamma_{r,c}$. As we see in some of the case the TE-CDE model does not even train completely due to early stopping criteria imposed by the authors. Further, in \cref{fig:fig_1} we show the training loss curves for various combinations of $\kappa$ and $\gamma_{r,c}$. Clearly, we see the disadvantage of using neural-controlled differential equations as they are unstable to train and diverge. Neural ODEs are well known for slower training as well as poor convergence properties. The neural-controlled differential equations also suffer from such issues and this is concerning since they are used in critical applications like counterfactual outcome estimation. Additionally, TE-CDE requires continuous values which are\ obtained through linear interpolation and this need not reflect the real-world data. On the other hand, S4model converges in every training setting with and outperforms TE-CDE by 100x at most. Their superior convergence makes them a reliable candidate for modelling longitudinal data. S4Model does not operate on continuous data hence it does not require any interpolation. This clear evidence suggests that indeed the general-purpose sequence models can beat the performance of continuous-time models like TE-CDE.

\begin{table*}[]
\centering
\caption{Results on test dataset. Here TC denotes Time Confounding parameters $\gamma_{r,c}$ and SI denotes sampling intensity $(\kappa)$. Here total loss in cross-entropy loss + RMSE}
\label{tab:result_1}
\resizebox{\textwidth}{!}{%
\begin{tabular}{@{}llllll|lllll@{}}
\toprule
Algorithm & \multicolumn{5}{c|}{TE-CDE}               & \multicolumn{5}{c}{S4Model}              \\ \midrule
Metric    & \multicolumn{5}{c|}{Total Loss} & \multicolumn{5}{c}{Total Loss} \\ \midrule
\multirow{2}{*}{\begin{tabular}[c]{@{}l@{}}TC ($\rightarrow$)\\ SI ($\downarrow$)\end{tabular}} &
  \multicolumn{1}{c}{\multirow{2}{*}{2}} &
  \multicolumn{1}{c}{\multirow{2}{*}{4}} &
  \multicolumn{1}{c}{\multirow{2}{*}{6}} &
  \multicolumn{1}{c}{\multirow{2}{*}{8}} &
  \multicolumn{1}{c|}{\multirow{2}{*}{10}} &
  \multicolumn{1}{c}{\multirow{2}{*}{2}} &
  \multicolumn{1}{c}{\multirow{2}{*}{4}} &
  \multicolumn{1}{c}{\multirow{2}{*}{6}} &
  \multicolumn{1}{c}{\multirow{2}{*}{8}} &
  \multicolumn{1}{c}{\multirow{2}{*}{10}} \\
 &
  \multicolumn{1}{c}{} &
  \multicolumn{1}{c}{} &
  \multicolumn{1}{c}{} &
  \multicolumn{1}{c}{} &
  \multicolumn{1}{c|}{} &
  \multicolumn{1}{c}{} &
  \multicolumn{1}{c}{} &
  \multicolumn{1}{c}{} &
  \multicolumn{1}{c}{} &
  \multicolumn{1}{c}{} \\ \midrule
$\kappa$ = 1     &12.2971        &3.5115        &20.7077        &31.8526        &11.3300       &\textbf{0.3407}        &\textbf{0.7396}        &\textbf{1.2541}        &\textbf{1.3585}       &\textbf{1.3773 }      \\
$\kappa$ = 5     &3.4005        &51.0468        & 3.4945       &4.4499        &38.0621      &\textbf{0.3358 }       &\textbf{0.7348}        &\textbf{1.2421}        &\textbf{1.3577}       &\textbf{1.3694}       \\
$\kappa$ = 10    &2.8759        &4.0433        &41.5988        & 10.3160       &22.8966       &\textbf{0.3377 }       &\textbf{0.7428}        &\textbf{1.2540 }       &\textbf{1.3609}       & \textbf{1.3726}      \\
$\kappa$ = 15    &59.9164        &3.2268        & 13.2311      &13.2911        &10.2217       &\textbf{0.3403}        &\textbf{0.7354 }       &\textbf{1.2465}        &\textbf{1.3583}       &\textbf{1.3742}       \\
$\kappa$ = 20    &5.1836        & 3.1370       &31.1797       &8.3600         & 25.7174      &\textbf{0.3387}        &\textbf{0.7396}        &\textbf{1.2501}        &\textbf{1.3530}       &\textbf{1.3762}       \\ \bottomrule
\end{tabular}%
}
\end{table*}

\begin{table}[]
\centering
\caption{Hyper parameter table}
\label{tab:hypertab}
\resizebox{\columnwidth}{!}{%
\begin{tabular}{@{}ll|ll@{}}
\toprule
\multicolumn{2}{c|}{S4model}        & \multicolumn{2}{c}{TE-CDE}           \\ \midrule
\multicolumn{1}{c}{Hyper-parameter name} & \multicolumn{1}{c|}{Value} & \multicolumn{1}{c}{Hyper-parameter name} & \multicolumn{1}{c}{Value} \\ \midrule
No. Layers                 & 4      & No. Layers                 & NA      \\
Latent dimension           & 128    & Latent dimension           & 256     \\
Batch size                 & 32     & Batch size                 & 32      \\
Learning rate              & 0.0005 & Learning rate              & 0.00001 \\
Optimizer                  & Adam   & Optimizer                  & SGD     \\
Number of epochs           & 50     & Number of epochs           & 50      \\
Number of training samples & 10000  & Number of training samples & 10000   \\
Number of test samples     & 1000   & Number of test samples     & 1000    \\ \bottomrule
\end{tabular}%
}
\end{table}

\begin{figure*}[!ht]
   \includegraphics[width=\textwidth]{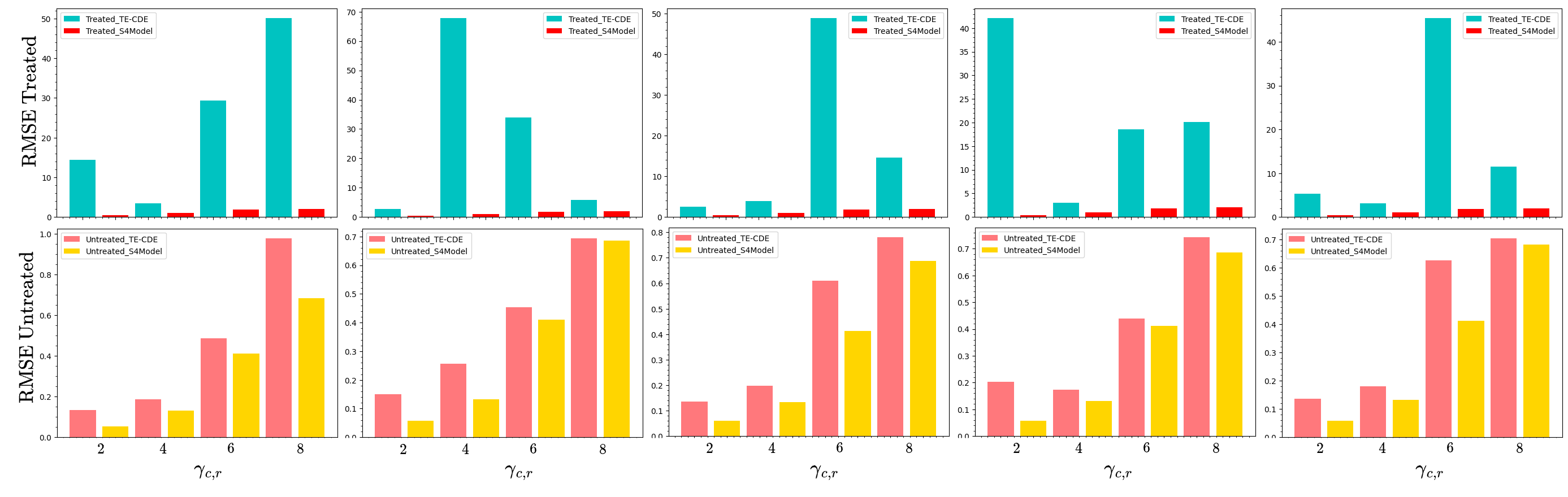}
   \vspace{-10mm}
   \caption{Bar plots showing treated v/s untreated RMSE calculated on the test dataset for TE-CDE and S4Model for various combination of $\kappa$ and $\gamma_{r,c}$}
   \label{fig:fig_2}
\end{figure*}
\subsection{Impact of Domain Adversarial Training}
As highlighted in \cite{tecde}, domain adversarial training has been shown to significantly improve the performance of TE-CDE. To investigate the effectiveness of this technique on our proposed method, S4Model, we conducted experiments to compare its performance with and without adversarial training. To do so, we used the loss function defined in \cref{eq:eq_7} and set $\mu = 0$ for both models. We evaluated the performance of S4Model and TE-CDE side-by-side on a dataset generated with $\kappa = 5$ for various levels of time-confounding. Our results show that the overall loss is significantly lower with domain adversarial training, and S4Model outperforms TE-CDE by a considerable margin. This highlights the effectiveness of domain adversarial training in achieving balanced representations, which can effectively remove bias introduced by time-dependent confounders and enable reliable counterfactual estimates. Our findings suggest that domain adversarial training is a useful technique for improving the performance of causal inference models, and should be considered when analyzing complex datasets.
\begin{table}[]
\centering
\caption{Adv v/s Non-Adv}
\label{tab:result_tavble_2}
\resizebox{\columnwidth}{!}{%
\begin{tabular}{@{}lll|ll@{}}
\toprule
Algorithm ($\rightarrow$) & \multicolumn{2}{c|}{TE-CDE}                            & \multicolumn{2}{c}{S4Model}                           \\ \midrule
Metric ($\rightarrow$)    & \multicolumn{2}{c|}{Total Loss}              & \multicolumn{2}{c}{Total Loss}              \\ \midrule
    Time-Confounding ($\downarrow$)  & \multicolumn{1}{c}{Balanced} & \multicolumn{1}{c|}{Imbalanced} & \multicolumn{1}{c}{Balanced} & \multicolumn{1}{c}{Imbalanced} \\ \midrule
$\gamma_{c,r}$ = 2 &3.4005   &18.0014  &\textbf{0.3358}  &1.8282  \\
$\gamma_{c,r}$ = 4     &51.0468 &3.3101  &\textbf{0.7348}  &1.9711  \\
$\gamma_{c,r}$ = 6   &3.4945  &5.8647  &\textbf{1.2421 } &2.9998  \\
$\gamma_{c,r}$ = 8      &4.4499  &4.8950  &\textbf{1.3577}  &2.9572  \\
$\gamma_{c,r}$ = 10    &38.0621  &11.2443  &\textbf{1.3694}  &2.7580  \\ \bottomrule

\end{tabular}%
}
\end{table}
\vspace{-3mm}

\begin{table}[]
\centering
\caption{ VRAM, time/epoch and number parameters of TE-CDE and S4Model}
\label{tab:vramtable}
\resizebox{\columnwidth}{!}{%
\begin{tabular}{@{}llll@{}}
\toprule
Algorithm & Training VRAM & Time/epoch(seconds) & \# of parameters \\ \midrule
TE-CDE    & 1037MiB              &   7.2                   & 2238149 \\
S4Model   & 1475MiB        &   4.4                 & 679487 \\ \bottomrule
\end{tabular}%
}
\end{table}

\begin{table}[]
\centering
\caption{Results of ablation study on number of layers in S4Model}
\label{tab:table_4}
\resizebox{\columnwidth}{!}{%
\begin{tabular}{@{}llll@{}}
\toprule
Number of layers & Treatment Loss & Outcome Prediction Loss & Total Loss \\ \midrule
2                &0.0240                & 0.3203                &0.3443          \\
4                &0.0267                & 0.3178                & 0.3445           \\
6                & 0.0257               & 0.3042                & 0.3299           \\
8                & 0.0240               & 0.2967                &  0.3207          \\
10               &0.0257                & 0.2956                 & 0.3213           \\ \bottomrule

\end{tabular}%
}
\end{table}
\subsection{Training Efficiency and Stability}
Apart from the superior empirical performance of S4Model we examine the several practical metrics that are important for real-world use. To this end we examine the amount of VRAM consumption, time taken per epoch, the total number of trainable parameters and inference speed. It is critical that the deployed model is smaller in memory footprint as well as faster on inferring from observations without compromising on the overall performance. In \cref{tab:vramtable} we show the necessary metrics. This shows that S4Model has 10x fewer parameters and is 100x faster in training as well as inference yet achieves 10x and in some cases 100x better performance than TE-CDE. As noted previously TE-CDE model can diverge during training. Even after multiple restarts, the training did not converge due to extremely large loss values leading to 'nan' errors.  Note that we re-run TE-CDE multiple times yet it is very sensitive to initialization conditions. Hence, making TE-CDE quite impractical for real-world use. 

\section{Abalation Study}
We study the effect of increasing the number of layers in S4Model and evaluate it's performance. Generally, increasing the number of layers should give us good performance improvements. We show the results in \cref{tab:table_4} total loss value, treatment loss and outcome prediction loss using various S4Models with layers {2, 4, 6, 8, 10}. We observe that increasing the number of layers decreases the overall loss values and does not overfit the training data. Another ablation study examines the latent dimension used by S4Model. We examine the empirical performance for the latent dimension of sizes {8, 16, 32, 128, 256} and show the results in \cref{tab:table_5}. We observe that S4Model is more expressive when the latent dimension is higher and benefits the performance.
\begin{table}[]
\centering
\caption{Results on ablation study of the latent dimension of S4Model}
\label{tab:table_5}
\resizebox{\columnwidth}{!}{%
\begin{tabular}{@{}llll@{}}
\toprule
Latent dimension & Treatment Loss & Outcome Prediction Loss & Total Loss \\ \midrule
8                &0.0528                &0.7607                 &  0.8135            \\
16               &0.0343               & 0.6842                &   0.7185         \\
32               &  0.0298               &0.5447                 &   0.5745         \\
128              & 0.0241               & 0.3748                &    0.3989        \\
256              & 0.0242               &  0.3181                &  0.3423          \\ \bottomrule
\end{tabular}%
}
\end{table}

\section{Conclusion and Future Scope}
In this paper, we examined the use of the general purpose sequence model - Structured state space model for counterfactual outcome estimation. We compared our method against the recently proposed method TE-CDE. Empirically our method showed superior results compared to TE-CDE on the lung cancer dataset. Under various time-confounding and sampling intensity settings we outperform TE-CDE with a lower memory footprint, 10x less number of parameters and guaranteed convergence. In summary, this work highlights the empirical performance, scalability and efficiency of state space models for counterfactual outcome prediction. As part of future work, we would like to examine the applicability of the state space model under a sparse data setting. We look for inspiration from compressive sensing theory \cite{dcs} and sparse coding to effectively reconstruct the sequence of observation from a small subset of observations.

\bibliography{example_paper}
\bibliographystyle{icml2023}



\end{document}